\documentclass[10pt,twocolumn,letterpaper]{article}

\usepackage[pagenumbers]{wacv} 

\usepackage{amsmath,amssymb,mathtools}
\usepackage{booktabs}
\usepackage{graphicx}
\usepackage{xspace}
\usepackage{xcolor}
\usepackage{enumitem}
\usepackage{url}
\usepackage{amsthm}

\definecolor{wacvblue}{rgb}{0.21,0.49,0.74}
\usepackage[pagebackref,breaklinks,colorlinks,allcolors=wacvblue]{hyperref}

\def\wacvPaperID{2745} 
\def\confName{WACV}
\def\confYear{2027}

\title{Adapting Without Gradients:\\Affine Statistics Transport and What Its Certificate Can Tell You}

\author{Salim Khazem\\
Talan Research Center\\
Paris, France\\
{\tt\small salim.khazem@talan.com}
\and
Ibrahim Mohamed Serouis\\
Talan Research Center\\
Toulouse, France\\
{\tt\small ibrahim.mohamed-serouis@talan.com}
}

\begin{document}
\maketitle
\begin{abstract}
Test-time adaptation (TTA) typically assumes that model parameters can be updated at inference time. This assumption is restrictive for inference-only accelerators, frozen or third-party models, and memory-constrained deployments, and standard BatchNorm-based TTA configurations may also become inactive on architectures without BatchNorm. We study adaptation when the learned model must remain frozen. We introduce CASTER, a gradient-free method that stores source class statistics in a discriminative subspace, estimates a class-shared affine transformation from target-batch moments, and analytically transports the source class distributions before classification. CASTER requires no backward pass, optimizer state, or stored source feature bank. Across four backbones and seven datasets, it outperforms $k$-NN on identical frozen features in 27 of 28 backbone-dataset settings while retaining a median of $18\times$ less state. Affine transport is not always reliable. On ImageNet-C, where batches contain only 64 samples for 1000 classes, unconditional transport loses 21.2 top-1 points. We therefore introduce an empirical residual-to-margin transportability certificate. Across 307 evaluation cells, every transport losing more than 10 points has certificate value above 3.9, although benign and destructive regimes are not perfectly separated. Gating converts an average $-3.35$-point effect of unconditional transport into a $+1.69$-point gain, and performance remains within 0.3 points of the best threshold over a broad threshold range. Finally, we show that this certificate is mechanism-specific: when applied to Tent, it accepts only $4.3\%$ of updates and preserves $0.6\%$ of Tent's available gain. These results position CASTER as a lightweight adaptation mechanism for frozen-model deployment, together with an explicit account of when its safety signal is informative and when it is not. Code is available in \url{https://github.com/salimkhazem/Caster.git}
\end{abstract}

\section{Introduction}
\label{sec:intro}

\begin{figure}[t]
    \centering
    \includegraphics[width=\columnwidth]{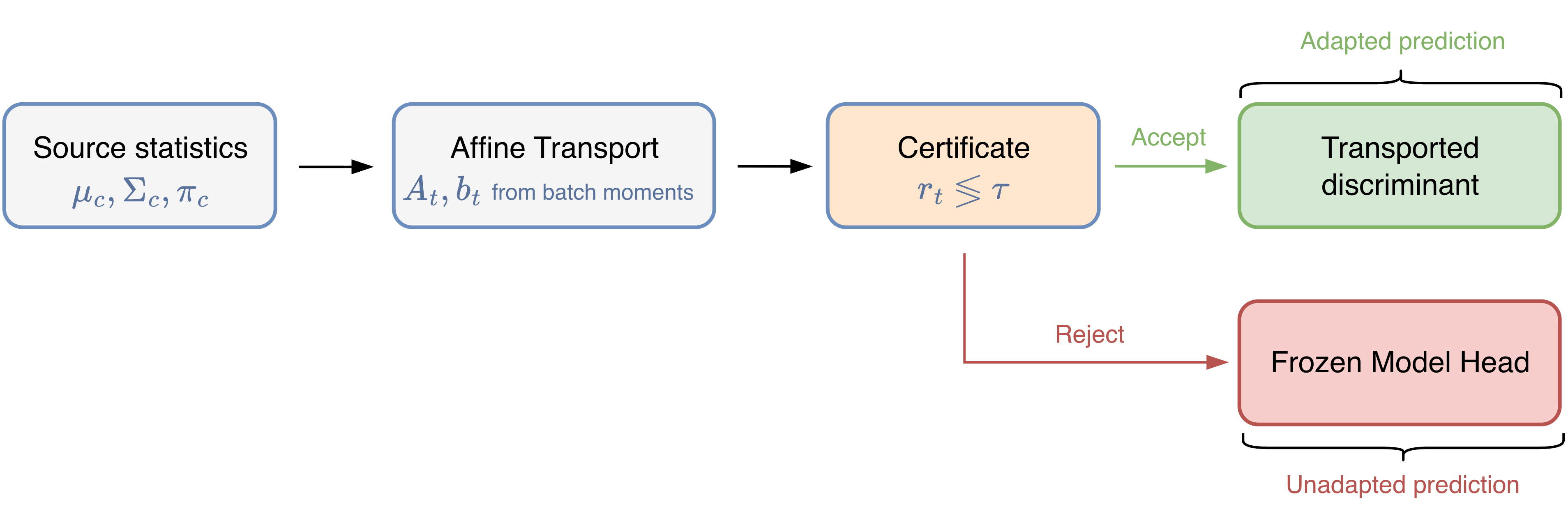}
    \caption{\textbf{CASTER at test time.} Source class moments are transported to the target batch by a class-shared affine map estimated from batch statistics. The transported Gaussian discriminant is used only when the transportability certificate passes; otherwise CASTER returns the frozen model-head prediction. No gradients are computed and no source feature bank is stored.}
    \label{fig:method}
\end{figure}

Test-time adaptation (TTA) aims to recover accuracy under distribution shift by using unlabeled target data available only at inference time. Strong methods such as Tent~\cite{wang2021tent}, EATA~\cite{niu2022eata}, CoTTA~\cite{wang2022cotta}, and SAR~\cite{niu2023sar} update model parameters online and can be highly effective when a backward pass and mutable model state are available. In many deployments, however, those assumptions are restrictive: inference-only accelerators may not expose backward computation, frozen binaries or external APIs may not expose trainable parameters, and optimizer state can violate latency or memory budgets. Forward-only TTA such as FOA~\cite{niu2024foa} relaxes the backward-pass requirement through derivative-free prompt optimization, but still performs a test-time search. We study a stricter regime in which the learned network parameters remain frozen and adaptation is performed analytically in feature space.

A second practical issue is architectural. In the standard implementations evaluated here, Tent and EATA collect BatchNorm affine parameters. Consequently, on our four LayerNorm backbones ConvNeXt-T, ViT-B/16, DeiT-B, and Swin-T their eligible parameter set is empty and their predictions reduce exactly to the frozen source model. SAR differs because it can update BatchNorm, GroupNorm, or LayerNorm parameters. This distinction matters when reporting TTA results across modern architectures: an average that mixes active and inert configurations can obscure what a method is actually adapting.

We propose \textbf{CASTER}, a gradient-free TTA method based on \emph{class-shared affine statistics transport}. Our hypothesis is that many practically relevant shifts approximately transform a source-trained feature space by a single affine map shared across classes. CASTER stores only source class moments in a discriminative subspace. For each target batch, it estimates a shrinkage-stabilized map from source and target batch moments, transports the source class Gaussians through that map, and evaluates the resulting Gaussian discriminant. The feature extractor and learned network parameters are never updated, no source feature bank is retained, and no optimizer state is required.

Analytic transport is not unconditionally safe. Its weakest regime is also easy to diagnose: when the number of classes is large relative to the target batch, pseudo-class coverage is too sparse to support reliable class geometry. On ImageNet-C, for example, a batch contains $64$ samples for $1000$ classes, and unconditional transport loses $21.2$ top-1 points relative to the frozen head. CASTER therefore includes a \emph{transportability certificate}: after inverse transport, confident pseudo-class centroids must remain small relative to the minimum source-class margin, measured in the same Mahalanobis geometry.

A central result of this paper is a precise characterization of what that certificate does and does not justify. It reliably identifies the catastrophic transport regime: every admissible evaluation cell that loses more than $10$ points has certificate value above $3.9$. The induced policy is also insensitive to the exact threshold; across a roughly fourfold interval, performance stays within $0.3$ points of the best threshold. However, the raw certificate score does \emph{not} rank adaptation benefit within the non-degenerate regime: its correlation with the gain from transport is positive, whereas a score used to prefer beneficial updates would need the opposite ordering. We therefore present the certificate as a safety floor for the transport mechanism, not as a general predictor of how much adaptation will help.

That distinction becomes sharper when the same gate is applied to a different adaptation mechanism. On matched CIFAR-10-C cells, the certificate admits only $4.3\%$ of Tent updates and preserves just $0.6\%$ of Tent's available gain; a more permissive source-statistic referee accepts $46.3\%$ and still keeps only $32.3\%$. At the cell level, the certificate's acceptance rate decreases as Tent's ungated gain increases ($\rho=-0.86$), whereas the analogous relationship is positive for CASTER's own transport. The failure is structural: a criterion computed from the pre-update frozen representation cannot, in general, certify the benefit of a mechanism that changes that representation.

\noindent\textbf{Contributions.}
We make three contributions. \textbf{(i)} We introduce CASTER, an analytic, gradient-free TTA mechanism that stores per-class moments rather than a source feature bank. On identical frozen features it outperforms $k$-NN in $27$ of $28$ backbone-dataset cells, with an $18\times$ smaller median retained state, and it remains applicable to architectures on which the BatchNorm-only Tent/EATA configurations evaluated here are inert. \textbf{(ii)} We derive a scale-consistent residual-to-margin certificate and empirically separate two claims that are often conflated: detecting degenerate transport and ranking adaptation benefit. The certificate supports the former, not the latter. \textbf{(iii)} We show that this certificate does not transfer to a feature-updating mechanism, providing both matched experiments and a structural explanation for the failure. All reported aggregates are regenerated from raw logs by a fail-closed verification script to reduce silent evaluation errors.

\section{Related Work}
\label{sec:related}

\paragraph{Gradient-based test-time adaptation.}
Tent minimizes prediction entropy by updating normalization affine parameters and test-time normalization statistics~\cite{wang2021tent}. EATA adds reliability and redundancy filtering together with an anti-forgetting regularizer~\cite{niu2022eata}; CoTTA uses teacher-style averaging, augmentation, and stochastic restoration for non-stationary streams~\cite{wang2022cotta}; and SAR combines reliable sample selection with sharpness-aware optimization to improve stability under small batches and dynamic shifts~\cite{niu2023sar}. These methods can be strong when backward computation and mutable parameters are available. In our experiments we report their advantage over CASTER where it exists rather than restricting comparisons to settings favorable to frozen-feature methods (Sec.~\ref{sec:corruption}).

\paragraph{Backward-free and frozen-model adaptation.}
T3A replaces the linear classifier by pseudo-prototypes formed from confident target samples while keeping the feature extractor fixed~\cite{iwasawa2021t3a}. FOA removes backpropagation by optimizing an added input prompt with a derivative-free evolution strategy and by shifting activations toward source statistics~\cite{niu2024foa}. Recent work has also explored parameter-efficient adaptation of frozen vision backbones, including low-rank adaptation strategies for Vision Transformers~\cite{khazem2026adaptertune} and topology-aware LoRA adaptation for segmentation~\cite{khazem2026topolora}. These approaches retain a largely frozen pretrained representation while introducing a small set of trainable parameters. They nevertheless require parameter updates during adaptation. CASTER considers a stricter deployment regime in which the learned network remains entirely frozen and adaptation is performed analytically in feature space. In particular, CASTER estimates a closed-form affine transport of source class statistics from each target batch, without backpropagation, trainable parameters, or test-time optimization. We also compare with $k$-NN over the same frozen features as a deliberately strong nonparametric estimator; unlike CASTER, its retained state grows with the number of source examples.

\paragraph{Selective adaptation and risk monitoring.}
Selective prediction and learning-to-defer study whether a model should act when a confidence criterion is satisfied. Relatedly, recent work monitors predictive risk during TTA and raises an alert when a performance requirement is likely to be violated~\cite{schirmer_monitoring_2025}. Our question is narrower and mechanism-specific: before using an analytic transport, can we detect batches for which that transport is geometrically unsupported? The experiments in Secs.~\ref{sec:certificate} and~\ref{sec:specificity} show why this scope matters. A criterion validated for one adaptation mechanism should not be interpreted as a generic certificate for another.

\paragraph{Robustness benchmarks and distribution distances.}
CIFAR-C and ImageNet-C evaluate common corruptions~\cite{hendrycks2019robustness}; ImageNet-V2, ImageNet-R, and ImageNet-Sketch test natural or rendition-based shifts~\cite{recht2019imagenetv2,hendrycks2021many,wang2019sketch}; and PACS and Office-Home are standard domain-shift benchmarks~\cite{li2017deeper,venkateswara2017officehome}. Distribution distances such as maximum mean discrepancy (MMD)~\cite{gretton2012mmd} and Wasserstein distance~\cite{villani2009optimal} quantify how far two distributions are apart. Distance from the source, however, is not the same as the expected benefit of a particular adaptation rule. This distinction is central to our certificate analysis.

\section{Method}
\label{sec:method}

Let $h=f_\theta(x)\in\mathbb{R}^d$ denote the output of a frozen feature extractor and let $y\in\{1,\ldots,C\}$. CASTER performs all adaptation outside the learned network. From labeled source data, we first construct a discriminative linear subspace from the generalized eigenvectors of the between-class and within-class scatter matrices. Let $U_{k_0}\in\mathbb{R}^{d\times k_0}$ contain the leading directions, where
\begin{equation}
    k_0=\min(k_{\max},\,C-1).
\end{equation}
The $C-1$ bound reflects the maximum rank of between-class scatter. At target time, for a batch of size $B$, we activate the first
\begin{equation}
    k=\min(k_0,\,B-1)
    \label{eq:kchoice}
\end{equation}
directions and use $z=U_k^\top h$. The $B-1$ bound avoids carrying directions that cannot be supported by the empirical target covariance before regularization. Source statistics are stored once in the largest subspace $U_{k_0}$ and restricted to the first $k$ coordinates as needed. Thus the retained source state does not depend on the number of source examples.

For each class $c$, CASTER stores the projected source mean $\mu_c\in\mathbb{R}^k$, covariance $\Sigma_c\in\mathbb{R}^{k\times k}$, and prior $\pi_c$. Let $m_s$ and $C_s$ denote the corresponding global source mean and covariance in the active subspace.

\subsection{Affine Statistics Transport}
\label{sec:transport}

For a target batch $\mathcal{B}_t=\{z_i\}_{i=1}^{B}$, let $\hat m_t$ and $\hat C_t$ be its empirical mean and covariance. CASTER estimates a shrinkage-stabilized whiten-color map
\begin{equation}
\begin{aligned}
A_t &= (\hat C_t+\lambda I)^{1/2}(C_s+\lambda I)^{-1/2},\\
b_t &= \hat m_t-A_t m_s,
\end{aligned}
\label{eq:transport}
\end{equation}
where matrix square roots are the symmetric positive-semidefinite roots and $\lambda>0$ controls covariance shrinkage. When $B\ge 2k$, we use the full matrix in Eq.~\eqref{eq:transport}. When $B<2k$, we use its diagonal analogue, replacing the two covariance matrices by their diagonal variances. This rule is a sample-efficiency safeguard: the dense map has $\mathcal{O}(k^2)$ degrees of freedom, whereas the diagonal map has $\mathcal{O}(k)$.

The source class statistics are transported as
\begin{equation}
    \tilde\mu_{c,t}=A_t\mu_c+b_t,
    \qquad
    \tilde\Sigma_{c,t}=A_t\Sigma_cA_t^\top+\beta I,
\end{equation}
with $\beta>0$ providing class-covariance regularization. The transported Gaussian discriminant score is
\begin{equation}
\begin{aligned}
g_c(z)=&-\tfrac{1}{2}(z-\tilde\mu_{c,t})^\top
\tilde\Sigma_{c,t}^{-1}(z-\tilde\mu_{c,t})\\
&-\tfrac{1}{2}\log\det(\tilde\Sigma_{c,t})+\log\pi_c.
\end{aligned}
\label{eq:gda}
\end{equation}
The proposed target prediction is $\hat y_i=\arg\max_c g_c(z_i)$. No network parameter is changed by this operation.

\subsection{Transportability Certificate}
\label{sec:certificate_method}

The affine map is only useful when a target batch contains enough evidence to support the transported class geometry. CASTER therefore checks the proposed transport before committing to its predictions.

We retain target samples whose transported-GDA confidence exceeds a fixed threshold and let $I_{c,t}$ be the retained samples assigned to class $c$. Each retained feature is inverse-transported to source coordinates,
\begin{equation}
    \bar z_i=A_t^{-1}(z_i-b_t),
\end{equation}
and for every occupied pseudo-class $c$ we compute the centroid residual
\begin{equation}
    \hat\delta_{c,t}
    =\frac{1}{|I_{c,t}|}\sum_{i\in I_{c,t}}\bar z_i-\mu_c.
\end{equation}
Let $\Sigma_s$ be the pooled source within-class covariance and define the minimum source-class separation in that same metric as
\begin{equation}
    \gamma_s
    =\min_{c\neq c'}
    \left[(\mu_c-\mu_{c'})^\top\Sigma_s^{-1}
    (\mu_c-\mu_{c'})\right]^{1/2}.
\end{equation}
The certificate is
\begin{equation}
    r_t=
    \frac{
    \max_{c:\,|I_{c,t}|>0}
    \left(\hat\delta_{c,t}^\top\Sigma_s^{-1}\hat\delta_{c,t}\right)^{1/2}
    }{\gamma_s+\varepsilon}.
\label{eq:certificate}
\end{equation}
CASTER accepts the transported classifier only if $r_t\le\tau$ and the retained set satisfies fixed minimum-sample and minimum-class-coverage requirements. If either coverage requirement fails, or no confident pseudo-class remains, the batch is rejected by construction and the frozen model head is used instead. All thresholds are fixed globally rather than tuned per dataset.

Using the same Mahalanobis metric in the numerator and denominator is essential. If a Euclidean centroid residual is divided by a Mahalanobis class margin, the ratio changes under an invertible rescaling of feature coordinates and therefore ceases to measure transport quality independently of feature conditioning. Equation~\eqref{eq:certificate} is dimensionless and invariant to such linear rescalings. Empirically, this correction is also necessary: on Flowers-102, the mixed-metric version can remain numerically small even when an inverse-transported centroid approaches a different source class mean, making any useful rejection threshold impossible. Section~\ref{sec:certificate} evaluates the resulting gate and, importantly, separates catastrophe detection from benefit ranking.

\section{Experiments}
\label{sec:experiments}

\subsection{Protocol}

\paragraph{Frozen-feature estimator quality.}
We first isolate the quality and storage cost of the classifier used on a fixed representation. We evaluate four ImageNet-pretrained backbones (ViT-B/16, ResNet-50, DeiT-B, and Swin-T) on seven datasets (Oxford-IIIT Pets, DTD, Flowers-102, Food-101, CIFAR-10, CIFAR-100, and Tiny-ImageNet). The backbones remain frozen and receive no task-specific fine-tuning. Every estimator consumes the same extracted features, so accuracy differences are attributable to the estimator rather than representation learning. We compare CASTER's source Gaussian discriminant with $k$-NN and report the retained state required at inference: the complete source feature bank for $k$-NN versus per-class moments for CASTER.

\paragraph{Corruption benchmarks.}
For CIFAR-10-C and CIFAR-100-C we evaluate all $19$ corruptions at all $5$ severities over six backbones. Source checkpoints are trained for the corresponding clean task. A run is admitted only if (i) the checkpoint exists, (ii) clean-source accuracy passes a sanity threshold (CIFAR-10 $\ge85\%$, CIFAR-100 $\ge55\%$), and (iii) the source moments were generated from that exact checkpoint. ResNet-50 is repeated over three seeds; deterministic or single-seed backbones are reported without inventing seed replication. Our verification script fails closed when a required artifact is missing or inconsistent.

\paragraph{Measurement conventions.}
Two conventions are important. First, harm is always measured against an immutable copy of the unadapted source model. Comparing an in-place adaptive model with its own immediately preceding state would make harmful updates appear artificially benign. Second, when an adaptation baseline has no eligible parameters, its prediction must reproduce source-only inference exactly, including evaluation mode. Otherwise dropout or other train-mode behavior can manufacture an apparent adaptation effect.

\subsection{Frozen-Feature Estimation}
\label{sec:estimator}

\begin{figure}[t]
    \centering
    \includegraphics[width=\columnwidth]{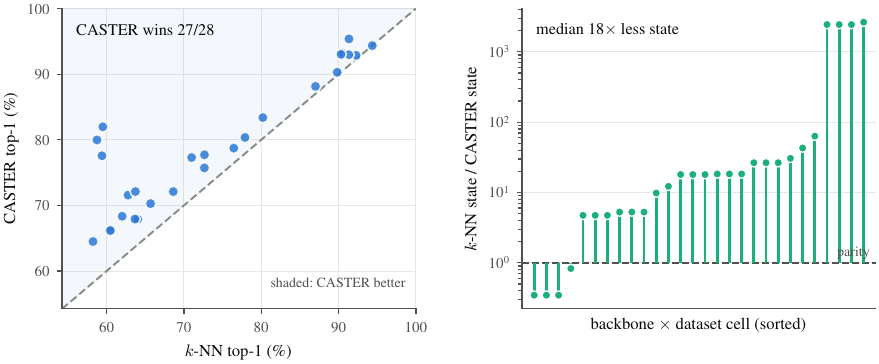}
    \caption{\textbf{Frozen-feature estimator comparison} over four backbones and seven datasets. All methods use identical frozen ImageNet-pretrained features. \textbf{Left:} CASTER versus $k$-NN top-1 accuracy; points above the diagonal favor CASTER. \textbf{Right:} retained $k$-NN state relative to CASTER, on a log scale. Cells where the ratio falls below one are reported explicitly in the text.}
    \label{fig:estimator}
\end{figure}

Figure~\ref{fig:estimator} compares CASTER's source Gaussian estimator with $k$-NN on identical frozen features. CASTER is better in $27$ of $28$ backbone--dataset cells. The only exception is ViT-B/16 on CIFAR-10, where the two methods differ by less than $0.04$ top-1 points. The largest margins occur on Flowers-102: CASTER gains approximately $18$--$22$ points on three of the four backbones, a regime with many classes and relatively few labeled examples per class.

The right panel reports retained state. A $k$-NN bank grows linearly with the number of source examples and reaches $781$\,MiB on Tiny-ImageNet. CASTER stores class moments, with state $\mathcal{O}(Ck_0^2)$ and no dependence on the number of source examples. This advantage is not universal: on Flowers-102, only $1020$ training images are stored by $k$-NN while $k_0=C-1=101$ makes CASTER's covariance state comparatively large. We report that inversion rather than hiding it.

\subsection{What the Certificate Detects}
\label{sec:certificate}

A useful gate can support two very different claims: \emph{catastrophe detection} and \emph{benefit ranking}. Our certificate supports the first but not the second.

\begin{figure}[t]
    \centering
    \includegraphics[width=\columnwidth]{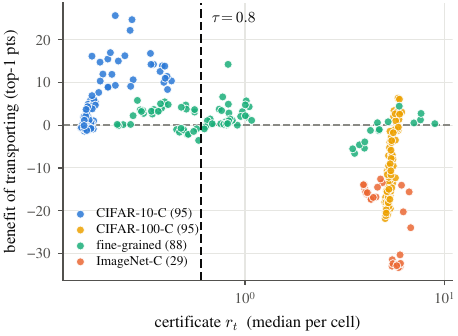}
    \caption{\textbf{Certificate score versus realized transport gain}, one point per admissible evaluation cell on i.i.d. streams. Every cell losing more than $10$ points lies above $3.9$, but the safe and unsafe bands are not disjoint. The symmetric-log axis makes both the benign and catastrophic regimes visible.}
    \label{fig:certificate}
\end{figure}

\paragraph{It detects degenerate transport.}
Figure~\ref{fig:certificate} pools all admissible evaluation cells from CIFAR-10-C, CIFAR-100-C, ImageNet-C, and the additional target datasets used in the transport study. Every cell that loses more than $10$ top-1 points under transport has certificate value above $3.9$. The catastrophic cells are concentrated in low-coverage regimes in which the class count is large relative to the target batch, most notably ImageNet-C with $1000$ classes and batch size $64$.

We do not claim perfect separation. The safe and unsafe bands overlap: at least one non-harmful cell has a certificate as high as $8.9$. This matters because an earlier analysis used fine-grained test streams that were inadvertently class ordered, leaving $97$--$100\%$ of many batches dominated by one class. After shuffling to the intended i.i.d. stream, those cells are benign and the certificate reflects the corrected protocol. All results below use the i.i.d. streams.

\paragraph{It does not rank benefit.}
Within the non-degenerate regime, the raw certificate score is positively correlated with the accuracy gain from transport. A gate intended to rank updates by expected benefit would need the opposite ordering, because lower $r_t$ is the acceptance direction. The explanation is intuitive: larger distribution shifts tend to increase both the residual score and the amount of accuracy the degraded source head can recover. Consequently, on CIFAR-10-C, where catastrophic transport is absent, thresholding the certificate does not beat simply applying transport to every batch. The certificate should therefore be interpreted as a detector of unsupported transport, not as a utility score.

\paragraph{The threshold is not delicate.}
Across the pooled admissible cells, unconditional transport averages $-3.35$ points relative to the frozen head. Gating at the shipped $\tau=0.8$ gives $+1.69$ points; any $\tau\in[1.0,3.9]$ gives between $+1.85$ and $+1.92$ points. Thus a roughly fourfold threshold interval remains within $0.3$ points of the best observed value. We retain $\tau=0.8$ as a conservative operating point: it rejects some transports that would have been harmless rather than accepting transports in the catastrophic regime.

\begin{table}[t]
\centering\footnotesize
\setlength{\tabcolsep}{3.5pt}
\begin{tabular}{@{}l cc cc@{}}
\toprule
& \multicolumn{2}{c}{CIFAR-10-C} & \multicolumn{2}{c}{CIFAR-100-C} \\
\cmidrule(lr){2-3}\cmidrule(lr){4-5}
Method & BN & LN & BN & LN \\
\midrule
Frozen head & 72.0\,\tiny{$\pm$0.55} & 86.7\,\tiny{$\pm$0.82} & 48.4\,\tiny{$\pm$1.07} & 65.1\,\tiny{$\pm$0.61} \\
Source-GDA & 71.8\,\tiny{$\pm$0.64} & 86.3\,\tiny{$\pm$0.69} & 35.4\,\tiny{$\pm$0.43} & 57.4\,\tiny{$\pm$0.50} \\
T3A & 66.5\,\tiny{$\pm$0.79} & 86.7\,\tiny{$\pm$0.73} & 18.6\,\tiny{$\pm$0.32} & 63.0\,\tiny{$\pm$0.67} \\
Tent & 85.7\,\tiny{$\pm$0.60} & 86.7\,\tiny{$\pm$0.82} & 62.9\,\tiny{$\pm$0.58} & 65.1\,\tiny{$\pm$0.61} \\
EATA & \textbf{85.8\,\tiny{$\pm$0.60}} & 86.7\,\tiny{$\pm$0.82} & 62.9\,\tiny{$\pm$0.58} & 65.1\,\tiny{$\pm$0.61} \\
SAR & 85.7\,\tiny{$\pm$0.60} & 87.3\,\tiny{$\pm$0.74} & 62.9\,\tiny{$\pm$0.58} & 66.1\,\tiny{$\pm$0.85} \\
\textbf{CASTER} & 85.7\,\tiny{$\pm$0.51} & \textbf{87.4\,\tiny{$\pm$0.62}} & \textbf{63.4\,\tiny{$\pm$0.57}} & \textbf{68.1\,\tiny{$\pm$0.61}} \\
\bottomrule
\end{tabular}
\caption{\textbf{Corruption benchmarks}, aggregated over $19$ corruptions and $5$ severities and then grouped by normalization family (BN: ResNet-18/50; LN: ConvNeXt-T, ViT-B/16, DeiT-B, Swin-T). Entries are mean $\pm$ s.d. across backbones within each normalization family after averaging corruption/severity cells (and seeds where repeated) within each backbone. Tent and EATA use their standard BatchNorm-only parameter collection here and therefore reproduce the frozen head exactly on the four LN backbones.}
\label{tab:corruption}
\end{table}

\subsection{What the Gate Buys}
\label{sec:gate}

We next compare three policies on ResNet-50 across nine datasets: never transport, always transport, and transport only when the certificate accepts. No fixed policy dominates across regimes. Where transport helps, the gate gives up little: $0.1$ points on Flowers-102, $0.2$ on Oxford-IIIT Pets, and at most $2.2$ on CIFAR-10-C. Where transport is destructive, it recovers most or all of the loss: $+10.2$ points relative to always transporting on CIFAR-100-C and $+21.2$ on ImageNet-C, where every batch is rejected and CASTER therefore matches the frozen head.

Pooled over the admissible ResNet-50 cells, always transporting scores $55.70$ top-1 and never transporting scores $59.05$; certificate gating reaches $60.05$. The value of the gate is therefore asymmetric: rejecting a useful transport sacrifices at most the gain that transport could have provided, whereas accepting an unsupported transport can incur a much larger loss.

\subsection{Corruption Benchmarks and Normalization}
\label{sec:corruption}

Table~\ref{tab:corruption} separates BatchNorm and LayerNorm backbones. On the two BatchNorm ResNets, gradient-based adaptation is very strong, and CASTER does not claim a universal advantage. On the four LayerNorm architectures, the standard Tent and EATA implementations evaluated here collect no eligible BatchNorm parameters and therefore reduce exactly to the frozen source model. SAR remains active because it can adapt LayerNorm parameters. CASTER is independent of the network's normalization type because it operates on frozen extracted features rather than model parameters.

This comparison is intended to distinguish method capability from implementation artifacts. We do not claim that entropy minimization is conceptually impossible on LayerNorm networks; rather, the canonical BatchNorm-parameterized Tent/EATA configurations used in our benchmark are inactive there. Reporting normalization families separately prevents those inactive cells from being mistaken for evidence about an active adaptation mechanism.

\subsection{ImageNet-C}
\label{sec:imagenet}

ImageNet-C provides the clearest low-coverage stress test: there are $1000$ classes but only $64$ samples per target batch. The issue is not that a frozen representation is unusable in principle; it is that batch-level class geometry cannot be reliably inferred when most classes are absent from every batch.

Ungated feature-space alternatives fail sharply in this setting. Relative to the frozen head, Source-GDA and the non-transported Gaussian estimator lose $19.0$ points, unconditional transport loses $21.2$, and T3A loses $33.5$. The CASTER certificate rejects every target batch, so the deployed CASTER policy returns the frozen-head prediction and incurs zero additional harm. SAR is the only evaluated adaptive method that improves over the frozen head in this experiment ($+4.6$ points), consistent with its different mechanism of updating normalization parameters rather than estimating target class geometry.

This result illustrates the intended use of the certificate: when the assumptions required by affine class-statistics transport are not supported by the observed batch, CASTER abstains from that mechanism instead of forcing an update.

\subsection{Certificates Do Not Transfer Across Mechanisms}
\label{sec:specificity}

A mechanism-specific certificate should not automatically be interpreted as a generic test of whether \emph{any} adaptation is safe. We test this directly by placing two pre-update gates in front of Tent and EATA: CASTER's transportability certificate and an independent source-statistic referee based on agreement with source-derived predictions. The comparison uses $57$ matched CIFAR-10-C cells with ResNet-50 ($19$ corruptions $\times$ $3$ severities), paired with the corresponding ungated runs.

\begin{table}[t]
\centering
\small
\setlength{\tabcolsep}{3pt}
\resizebox{\linewidth}{!}{%
\begin{tabular}{l r r r r r}
\toprule
\textbf{Policy} & \textbf{Top-1} & \textbf{$\Delta$} & \textbf{Accept} & \textbf{Harm} & \textbf{Kept} \\
\midrule
Frozen head & 71.52 & +0.00 & --- & 0.00\% & --- \\
\midrule
\multicolumn{6}{l}{\emph{Certificate applied to the mechanism it was designed for}} \\
Transport, ungated & 86.18 & +14.66 & 100.0\% & 4.52\% & 100.0\% \\
\textbf{CASTER} (gated) & \textbf{87.50} & +15.98 & 91.0\% & 0.92\% & --- \\
\midrule
\multicolumn{6}{l}{\emph{The same pre-update criterion applied to Tent}} \\
Tent, ungated & 86.87 & +15.35 & 100.0\% & 2.97\% & 100.0\% \\
Tent + certificate gate & 71.61 & +0.09 & 4.3\% & 0.07\% & 0.6\% \\
Tent + referee gate & 76.47 & +4.95 & 46.3\% & 1.11\% & 32.3\% \\
\bottomrule
\end{tabular}%
}
\caption{\textbf{A transportability certificate does not transfer automatically to a different adaptation mechanism.} Results are over $57$ matched CIFAR-10-C cells with ResNet-50. For CASTER, gating both improves top-1 and reduces harm by rejecting destructive transports. Applied to Tent, the same certificate accepts only $4.3\%$ of updates and preserves $0.6\%$ of Tent's available gain. ``Kept'' is omitted for gated CASTER because rejecting harmful transports can make its gain exceed that of the ungated policy. EATA shows the same qualitative behavior as Tent (within $0.03$ top-1 points).}
\label{tab:specificity}
\end{table}

Table~\ref{tab:specificity} shows that the transportability certificate leaves Tent essentially indistinguishable from the frozen model. The source-statistic referee is more permissive, yet still discards roughly two thirds of Tent's available improvement. Neither criterion is a useful safety layer for Tent.

The ordering explains the failure. At the cell level, certificate acceptance rate is strongly anti-correlated with Tent's ungated gain ($\rho=-0.86$): the gate closes most often on the cells where entropy minimization helps most. For CASTER's own transport, the analogous acceptance-rate relationship is positive ($\rho=+0.72$). These are mechanism-specific relationships, not evidence that the raw certificate score is a general benefit ranker; Sec.~\ref{sec:certificate} shows that it is not.

The failure is structural rather than a threshold-tuning issue. CASTER's certificate evaluates geometry in the \emph{pre-update frozen representation}. Tent subsequently changes normalization parameters and therefore changes the representation on which the certificate was computed. A pre-update statistic that never evaluates the post-update representation has no general basis for ordering the benefit of that update. We therefore treat the certificate as licensing CASTER's affine transport only, and recommend reporting any adaptation certificate together with the mechanism for which it was derived and validated.

\subsection{Cost}
\label{sec:cost}

Table~\ref{tab:cost} measures throughput and peak memory on the same CIFAR-10-C cell (fog, severity 3, ResNet-50), running one method at a time on an otherwise idle device. CASTER processes $1076$ images/s, compared with $667$ for Tent, $649$ for EATA, and $386$ for SAR. Thus CASTER is $1.6\times$ faster than Tent/EATA and $2.8\times$ faster than SAR on this controlled benchmark. Its $1249$\,MiB peak is $22.7\%$ of the $5510$\,MiB used by the gradient-based methods, and it retains $90\%$ of the frozen model's throughput.

We report controlled measurements rather than timings extracted from the large sweep, whose jobs shared GPUs. Under contention, relative timings were substantially distorted. SAR remains the most expensive evaluated method because its sharpness-aware update requires two forward--backward evaluations per adaptation step.

\begin{table}[t]
\centering
\setlength{\tabcolsep}{3.5pt}
\begin{tabular}{@{}l r r c@{}}
\toprule
Method & img/s & Mem. & Backward \\
 & & (MiB) & pass \\
\midrule
Frozen head & 1194 & 865 & --- \\
T3A & 1133 & 918 & --- \\
Source-GDA & 1142 & 1249 & --- \\
\textbf{CASTER} & 1076 & 1249 & --- \\
EATA & 649 & 5510 & $\checkmark$ \\
Tent & 667 & 5510 & $\checkmark$ \\
SAR & 386 & 5510 & $\checkmark$ \\
\bottomrule
\end{tabular}
\caption{\textbf{Cost of adaptation} on CIFAR-10-C fog, severity 3, with ResNet-50. Each method is measured in isolation on an idle device. CASTER avoids backward computation and optimizer state, retaining $90\%$ of frozen-head throughput while using $4.4\times$ less peak memory than the gradient-based methods.}
\label{tab:cost}
\end{table}

\section{Discussion and Limitations}

\paragraph{Gradient-based TTA remains highly competitive where model updates are available.}
On the BatchNorm backbones, gradient-based adaptation is strong, but the comparison is not uniformly in its favour: EATA is marginally higher than CASTER on the CIFAR-10-C BatchNorm aggregate, whereas CASTER is higher on the CIFAR-100-C BatchNorm aggregate. We therefore do not claim a universal advantage over gradient-based TTA. CASTER instead targets deployments in which learned model parameters must remain frozen and backward computation or optimizer state is unavailable, as well as architectures on which the BatchNorm-only Tent/EATA configurations evaluated here have no eligible parameters. SAR and other methods that adapt LayerNorm remain applicable in some of these settings, but require a different adaptation mechanism and, in the case of gradient-based methods, backward computation. More broadly, efficient visual learning increasingly requires balancing predictive performance with the computational and memory constraints of deployment~\cite{khazem2025polygonet}. CASTER follows this deployment-oriented perspective from a different angle: instead of compressing or modifying the learned representation, it keeps the representation fixed and minimizes the additional state and computation required for test-time adaptation.

\paragraph{The method inherits its representation.}
CASTER operates on a frozen feature space and cannot repair a representation that does not already separate the target classes sufficiently well. If the backbone provides poor class separation under a given shift, transporting source statistics cannot recover information that is absent from the representation. This limitation is shared by other frozen-feature estimators such as $k$-NN, which is why we compare estimators on identical extracted features rather than across independently trained pipelines.

\paragraph{Transport assumes a class-shared affine shift.}
CASTER assumes that the target feature distribution can be approximated by a single affine transformation shared across classes. This assumption can fail when the shift is strongly class-dependent or when the target batch is too small or imbalanced for its empirical moments to represent the target distribution reliably. In particular, when only a small subset of classes is present in a batch, variation in class composition can be confounded with distribution shift. The transportability certificate is designed to identify empirically unsupported transports in such regimes, but it is a diagnostic rather than a formal guarantee. We therefore characterise its behaviour empirically and make no claim of safety over arbitrary target distributions.

\paragraph{The certificate is mechanism-specific.}
The residual-to-margin statistic is derived from the geometry induced by CASTER's affine transport and should be interpreted only in that context. Its failure to gate Tent does not imply that feature-updating methods cannot be monitored; rather, it shows that a criterion computed solely from the pre-update frozen representation has no general guarantee of ranking the benefit of an update that subsequently changes that representation. Certificates for other adaptation mechanisms should therefore be derived and validated against the mechanisms they are intended to gate.

\section{Conclusion}
\label{sec:conclusion}

We studied test-time adaptation under frozen-model constraints, where learned network parameters cannot be modified and backward computation or optimizer state may be unavailable. CASTER transports source class statistics to each target batch through a class-shared affine map estimated from batch moments, requiring neither a backward pass nor a stored source feature bank. On identical frozen features, it outperforms $k$-nearest neighbours in 27 of 28 backbone--dataset settings while achieving a median $18\times$ reduction in retained state.

Affine transport is not unconditionally reliable. Across $307$ admissible evaluation cells spanning nine datasets, unconditional transport is worse on average than retaining the frozen head. CASTER therefore uses an empirical residual-to-margin transportability certificate to decide whether to apply the transported classifier or fall back to the frozen prediction. Unconditional transport changes accuracy by an average of $-3.35$ points relative to the frozen head, whereas gating at the shipped threshold yields $+1.69$ points. Moreover, a roughly fourfold range of thresholds remains within $0.3$ points of the best observed operating point. The certificate therefore provides a useful safety floor for affine transport, while our analysis also shows its limitation: it detects empirically degenerate transport but does not rank how much a non-degenerate adaptation will help.

Finally, we show that this transportability certificate is mechanism-specific. Applied to Tent, it accepts only $4.3\%$ of updates and preserves $0.6\%$ of Tent's available gain, demonstrating that evidence for a certificate on one adaptation mechanism should not be transferred automatically to another. Where model updates and backward computation are available, gradient-based TTA remains a strong alternative and can outperform CASTER in some regimes. CASTER instead provides a lightweight option for frozen-model deployment, together with an explicit account of when its adaptation mechanism is supported by the observed target batch and when it should abstain.

{
    \small
    \bibliographystyle{ieeenat_fullname}
    \bibliography{refs}
}

\end{document}